\documentclass[runningheads,a4paper]{llncs}

\usepackage{amssymb}
\usepackage{graphicx}
\usepackage{enumerate}
\usepackage{paralist}
\usepackage{hyperref}
\usepackage{url}

\usepackage[ruled,section]{algorithm}
\usepackage[noend]{algorithmic}
\usepackage{amssymb}
\usepackage{fca}
\usepackage{amsmath}
\usepackage{amsfonts}
\usepackage{alltt}

\usepackage{url}
\usepackage{amsmath}
\urldef{\mailsa}\path|chibe@edu.hse.ru|
\urldef{\mailsb}\path|emitrofanova@hse.ru|

\newcommand{\keywords}[1]{\par\addvspace\baselineskip
\noindent\keywordname\enspace\ignorespaces#1}

\begin{document}

\mainmatter  

\titlerunning{Boolean Matrix Factorisation for Collaborative Filtering: An FCA-Based Approach}
\authorrunning{Ibe Chukwuemeka Emmanuel \and Ekaterina Mitrofanova}

\title{Fairness of Machine Learning Algorithms in Demography}

\titlerunning{Fairness of Machine Learning Algorithms in Demography}

%
%
\author{Ibe Chukwuemeka Emmanuel \and Ekaterina Mitrofanova}
\authorrunning{}

\institute{National Research University Higher School of Economics, Moscow\\
\mailsa,\mailsb\\
}

%
%

\toctitle{Fairness of Machine Learning Algorithms in Demography}
\tocauthor{}
\maketitle

\begin{abstract}
The paper is devoted to the study of the model fairness and process fairness of the Russian demographic dataset by making predictions of divorce of the 1st marriage, religiosity, 1st employment and completion of education. Our goal was to make classifiers more equitable by reducing their reliance on sensitive features while increasing or at least maintaining their accuracy. We took inspiration from "dropout" techniques in neural-based approaches and suggested a model that uses "feature drop-out" to address process fairness. To evaluate a classifier's fairness and decide the sensitive features to eliminate, we used "LIME Explanations". This results in a pool of classifiers due to feature dropout whose ensemble has been shown to be less reliant on sensitive features and to have improved or no effect on accuracy. Our empirical study was performed on four families of classifiers (Logistic Regression, Random Forest, Bagging, and Adaboost) and carried out on real-life dataset (Russian demographic data derived from Generations and Gender Survey), and it showed that all of the models became less dependent on sensitive features (such as gender, breakup of the 1st partnership, 1st partnership, etc.) and showed improvements or no impact on accuracy




\end{abstract}

\section{Introduction}

Artificial Intelligence (AI) and Machine Learning (ML) are becoming more prevalent in various aspects of human life, especially those involving decision-making. Many of these algorithmic decisions are made without human supervision and by opaque decision-making processes. This raises questions about the possible bias of these processes towards certain groups of society, which may lead to unequal findings and potentially human rights abuses. One of the major challenges in maintaining public confidence is dealing with such biased models. 

Machine learning tasks often require the training of a model based on previous experience and data, which is then used for prediction and classification. Loan grants in the context of enacting legislation, detecting terrorism, predicting criminal recidivism, and other social and economic problems at a global level are examples of realistic applications where such models are used \cite{maes2002credit,addo2018credit,azizan2017terrorism}. These decisions have an effect on human life and can have negative consequences for society's most disadvantaged groups. The widespread use of machine learning algorithms has raised questions about user privacy, transparency, fairness, and trustworthiness. In order to make Europe ``fit for the digital age'', the European Union enacted the GDPR Law in 2016, which applies to all organisations and businesses. The law gives European citizens the right to have a clear understanding of how automated decision models function and to challenge their conclusions. The discriminatory automated decisions not only break anti-discrimination rules, but they also undermine public confidence in Artificial Intelligence. 

The following factors can contribute to undesirable bias in machine learning models:

\begin{itemize}
    \item The data collection \cite{roh2019survey} could be biased because some minority groups in society, such as people living in rural areas, do not produce sufficient data. Because of the unbalanced and biased datasets used in preparation, this results in an unfair model.
    \item If an incorrect model or training set is chosen, the training algorithm can be biased. Furthermore, when training, the model could take into account sensitive or discriminatory features, resulting in process unfairness.
\end{itemize}

Fairness has traditionally been associated with the results of decision-making processes \cite{SpeicherHGGSWZ18,ZafarVGG17}, with minor observation given to the process leading to the outcome \cite{Grgic-HlacaZGW18,grgic2016case}. These are based on the implementation of anti-discrimination laws in different countries of the world, which makes sure that individuals belonging to vulnerable groups (e.g., race, colour, gender, etc.) are treated equally.

This problem can be approached from a variety of perspectives, including:

\begin{itemize}
    \item Individual Fairness or Disparate Treatment \cite{SpeicherHGGSWZ18} takes into account people who belong to different sensitive groups but have similar non-sensitive characteristics and expects them to make the same decisions. For an example, during a job application process, job applicants with the same educational credentials should not be treated differently depending on their gender or race.
 \item Group Fairness or Disparate Impact \cite{SpeicherHGGSWZ18} notes that individuals from various sensitive attribute groups should achieve equal proportions of positive outcomes. To put it another way, it says, ``Different sensitive groups should be treated fairly''.

\item Disparate Mistreatment or Equal Opportunity \cite{ZafarVGG17} suggests that various sensitive groups should have comparable rates of decision-making error.

\item Process or procedural fairness \cite{Grgic-HlacaZGW18,grgic2016case} is concerned with the process that leads to the prediction and keeps track of the decision model's input features. In other words, process fairness is associated with the algorithmic level and makes sure that sensitive features are not used by the algorithm when making a prediction.

The difficulty of understanding machine learning models is a big issue when dealing with process fairness. Indeed, the black-box nature of ML models, such as deep neural networks and ensemble architectures like random forests (RF), makes it difficult to interpret and justify their outputs, and thus to trust their analyses by users and the general public. Explanatory models have been proposed to make Machine learning models more interpretable and transparent. Due to the complexities of recent Machine Learning models, asking for interpretations that might describe the model as a whole is unreasonable. As a result of this fact, local approaches to deriving possible explanations have emerged. The basic concept is to clarify the model at the local level rather than at the global level. 

\end{itemize}

The following desirable properties should be present in an ideal model explainer \cite{ribeiro2016should}:

\begin{itemize}
    \item Interpretability of the model: The model should allow for a contextual interpretation of the relationship between features and targets. The reasons should be simple and easy to understand.
  \item Local Fidelity: There is no possible reason that can be provided that explains the Machine learning results on every single instance. However, the explainer must be locally faithful to the instance being predicted at the very least.
  \item Model agnostic: The explainer should be capable of explaining any model.
  \item Global Perspective: The explainer should explain a representative sample to the user, such that the user has a global understanding of the explainer. LIME, Anchors, SHAP, and Deep Sift are examples of local explanatory methods \cite{ribeiro2016should,garreau2020explaining,ribeiro2018anchors,lundberg2017unified}. These are focused on ``linear explanatory approaches'', which have recently received a lot of attention due to their simplicity and applicability to a wide range of supervised machine learning scenarios. 
\end{itemize}

We use LIME (Local Interpretable Model Agnostic Explanations) in this research to extract local explanations for Machine learning classification models~\footnote{The exposition of theory and methods is based on~\cite{Alves:2020}}. LIME learns a surrogate linear model to approximate the Machine learning model in a neighbourhood around the target instance given a Machine learning model and a target instance. The coefficients of this linear model describe the feature contributions to the prediction of the target instance. As a result, LIME generates a list of the top features used by the Machine Learning model locally, as well as their contributions. 

For this study, $Lime_{global}$ was proposed as a method for deriving global explanations from LIME's locally important features. The  explanations can give us some insights into the process fairness of the model. This inevitably leads to the questions of how to ensure a fairer model in light of these explanations while minimising the effect on accuracy \cite{zafar2019fairness}. This prompted us to explore for $M_{final}$  models in which in comparison to the original model, their reliance on sensitive features is reduced and their accuracy is improved or at least maintained.

To attain both goals, a framework $Lime_{out}$  was proposed. This framework relies on feature dropout to produce a pool of classifiers that are then combined through an ensemble approach. Feature drop out receives a classifier and a feature $a$ as input, and then produces a classifier by ignoring $a$. Thus, feature $a$ is removed in both the training and the testing process.

The workflow of $Lime_{out}$ can be summarised as follows:  $Lime_{out}$ uses $Lime_{global}$ to evaluate the fairness of the given classifier by looking at the contribution of each feature to the classifier's outcomes, given the classifier provided by the user. The model is unfairly biased if the most important features include sensitive ones. Moreover, the model is thought to be unbiased. $Lime_{out}$ applies dropout of these sensitive features in the first instance, resulting in a pool of classifiers (as explained earlier). These are then combined to form $M_{final}$, an ensemble classifier. 

The paper is organised as follows. In Section ``Related Work'', we discuss some substantial papers explaining ability and fairness. We briefly observe LIME  for tabular data and discuss different fairness issues, some measures proposed in the literature, as well as the main motivation of the study. In Section ``Methodology'', we use LIME$_{out}$ to address issues of fairness based on the relative importance of features. In the Section ``Experiment'', we introduce the empirical dataset and indicate the feasibility of LIME$_{out}$. In Conclusion, we will consider some possible improvements to be accomplished in the future.

\section{Related work}

To obtain the interpretations of the Machine learning models, LIME \cite{ribeiro2016should} and Anchors \cite{ribeiro2018anchors} are frequently used. These methods provide the model with the most important features for predicting a particular instance. LIME and Anchors does not provide human-like reasons (they provide ``feature importance'' or contributions), and they have certain limitations \cite{garreau2020explaining}. In Section ``Methodology'', we use LIME to address issues of fairness based on the relative importance of features.

\subsection{LIME}
LIME (Local Interpretable Model Agnostic Explanations) is an interpretable surrogate linear model that mimics the action of a model locally. The feature space used by LIME does not have to be the same as model's feature space. A useful example of a representation by LIME is described in \cite{ribeiro2016should} as follows:
\begin{enumerate}

\item The binary vector representation of textual data that indicates the presence or absence of a word, and 

\item The binary vector which represents presence or absence of contiguous patch of similar pixels, in case of images.

\end{enumerate}

LIME can be described as follows \cite{ribeiro2016should}. Let $f:R^d \to R$  be the function learned by a classification or regression model over training samples. It is presumed that no supplementary knowledge about the function $f$ exists. Now consider $x \in R^d$ as an instance, and its prediction $f(x)$. LIME's aim is to clarify the $f(x)$ prediction locally. It should be noted that LIME's feature space does not have to be the same as $f$'s input space. In the case of text data, for example, interpretable space is represented by vectors that indicate if words are present or absent in a text, while the original space may be word embeddings or word2vec representations. As a matter of fact, LIME builds the local model using discretised features of smaller dimension d and aims to learn an explanatory model: $R^d \to R$ that approximates $f$ in the neighborhood of $x \in R^d$. LIME generates neighborhood points around an instance $x$ to be explained and assigns a weight vector to these points to obtain a local explanation.$\pi_x(z)$  which denotes the proximity measure of $z$ w.r.t. $x$, is being used to assign the weight. It then solves the following optimization problem to gain an understanding of the weighted linear surrogate model $g$:

$$g=argmin_{g\in G} L(f,g,\pi_x(z))+\Omega(g)$$

$L(f, g, \pi_x(z))$  is an estimate of how unreliable $g$ is at approximating $f$ in the locality interpreted by $\pi_x(z)$, and  is an estimate of $g$'s complexity (LIME uses the regularization term to measure complexity). LIME minimizes $L(f, g, \pi_x(z))$ thus enforcing $\Omega(g)$ to be small in order to be explainable by humans and also maintain both interpretability and local fidelity. The coefficients of $g$ correlate with each feature’s contribution to the prediction $f(x)$ of $x$.
LIME uses the following weighting function:    

$$\pi_x(z)=e^{\frac{D(x,z)^2}{\sigma^2}},$$

where $D(x,z)$ is the Euclidean distance between $x$ and $z$, and $\sigma$ is the hyper parameter (kernel-width). The value of $\sigma$ impacts the fidelity of explanation \cite{laugel2018defining}. When $\sigma$ is too large, for example, all instances are given equal weight, making it difficult to derive a linear model that will be able to explain all of them. If $\sigma$ is too small, only a couple of points are given significant weight, and even a constant model would be incapable to illustrate these points, resulting in lower coverage. As a result, we must choose the optimal $\sigma$ to guarantee coverage and local fidelity.

The workflow of LIME on tabular data \cite{garreau2020explaining} needs a training set (user defined) to generate neighborhood points. The following statistics are computed for each feature depending on their type:
\begin{itemize}

\item for categorical features it computes the frequency of each value;

\item for numerical features, it computes the mean and the standard deviation, which are then discretised into quartiles.
\end{itemize} 
 
Assume that $f$ is our feature model, and that we need to understand the prediction $f(\mathbf x)$ of $\mathbf x=(x_1, x_2, \ldots, x_n)$ where each  can be either a categorical or a numerical value. LabelEncoder is being used to map each categorical value to an integer. It should be noted that the values of each feature in the training set are divided into $p$ quantiles. The original instance is discretised using these quantile intervals. If $x_i$ is between $q_j$ and $q_{j+1}$, then the value $j$ is assigned. This is repeated for all features so as to obtain quantile boxes for all $x_i$. LIME samples discrete values from $\{1, \ldots, p\}$, $n$ times to obtain the perturbation $\hat{y}$ in the region of $\hat{x}$. LIME Tabular employs a normal distribution and quantile values to obtain the continuous representation $y$ of $\hat{y}$. The neighborhood instance $\hat{y}$ is expressed by a binary tuple, with the $i$-th component equal to 1 if $\hat{x}_l=\hat{y}_l$ and 0 if $\hat{x}_l\neq\hat{y}_l$. LIME Tabular generates all of these points in the neighborhood in this manner. The exponential kernel is used to assign weights to these points, and a weighted linear function is learned over the neighborhood permutations.

At the Figure 1, we represent the results of LIME explanation for religiosity of respondents. Blue colour represents the contributions of religious respondents, while orange colour represents the contributions of non-religious respondents.

\begin{figure}[H]
 \centering
\includegraphics[width=4.7in]{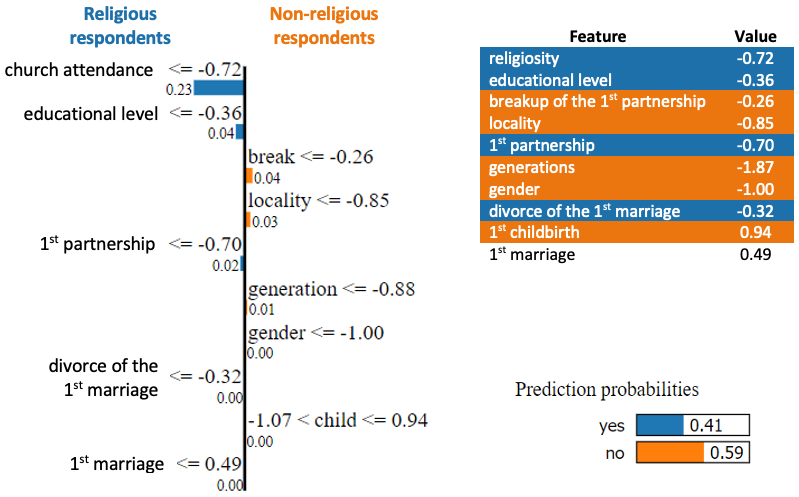}\\
  \caption{LIME explanation for religiosity}
  \label{fair_pic}
\end{figure}

\subsection{Model Fairness}

Several notions of model fairness based on decision results together with process fairness have been suggested \cite{dressel2018accuracy,SpeicherHGGSWZ18,ZafarVGG17,Grgic-HlacaZGW18,grgic2016case,Alves:2020}. Individual fairness \cite{chouldechova2017fair} also known as disparate treatment or predictive parity, ensures that instances or individuals belonging to different sensitive groups but with similar non-sensitive features must receive equal decision outcomes. The desire for different sensitive demographic groups to encounter equal rates of errors in decision outcomes is fundamental to the concept of group fairness also known as disparate impact or statistical parity \cite{dwork2012fairness}. COMPAS is a recidivism detection tool that uses a lengthy questionnaire to predict whether a prisoner would re-offend his crime. Northpointe (now Equivant), a commercial firm, created the famous algorithm at the time. 

According to a ProPublica8 study\footnote{\url{https://www.propublica.org/article/machine-bias-risk-assessments-in-criminal-sentencing}}, COMPAS has a significant ethnic bias. COMPAS is almost twice as likely to classify black people as high risk for non-re-offenders. Furthermore, white re-offenders are predicted as low risk even more often than black offenders in COMPAS. In other words, as opposed to white defendants, COMPAS has significantly higher false positive and lower true negative rates for black defendants. COMPAS is used by judges and parole officers across the United States to determine whether to grant or refuse probation to offenders; therefore, it is critical to understand how this model comes to its conclusions and ensure that it is fair. If we focus on the decision outcomes, the fair algorithm in the case of COMPAS (assuming Race is the only sensitive feature) should be as follows:

\begin{enumerate}

\item Black and whites with the same features get the same output (no disparate treatment and thus non-discriminatory)

\item The proportion of individuals classified as high-risk should be same across both the groups (statistical parity).

\end{enumerate}

We can deal with this bias during training \cite{ZafarVGG17} by:
\begin{enumerate}

\item Eliminating all possible features that may cause the model to create bias, e.g., race, gender etc.

\item Including discrimination measures as learning constraints, i.e., the model should be trained to minimize $P(y_{pred} \neq y_{true})$  such that
$$P(y_{pred} \neq y_{true} \mid race = Black)=P(y_{pred} \neq y_{true} \mid race=White),$$ 
where $y_{pred}$ is the risk predicted by trained Machine learning model (e.g., COMPAS) and $y_{true}$ is the true risk value.
\end{enumerate}

Given the fact that ``race'' is a sensitive feature motivates this limitation. When such restrictions are applied to various sensitive features separately (e.g., sex, ethnicity, nationality, etc.), it can result in unfairness for groups that are at the intersection of several types of discrimination (e.g., black women), a practice known as fairness gerrymandering \cite{kearns2018preventing}. Constraints for numerous combinations of sensitive attributes, on the other hand, make model training extremely difficult and can contribute to overfitting. Individual and group fairness have previously been considered as conflicting measures of fair ML \cite{zemel2013learning,zafar2019fairness}, and some studies have attempted to find an acceptable trade-off between them. The author argues in \cite{Binns20} that, despite their obvious contradictions, they all refer to the same basic moral concept. The author takes a wider view and argues for individual treatment and evaluation on a case-by-case basis. 

In \cite{Grgic-HlacaZGW18,grgic2016case}, the author discusses another important metric for measuring fairness, which is process fairness. In \cite{Grgic-HlacaZGW18} the author demonstrates a crucial insight into relying on a human's moral judgment or intuition about the fairness of using an input feature in algorithmic decision making. He also evaluates the impact of eliminating such input features on the classifier's accuracy, and designs an acceptable trade-off for the classifier's accuracy and process fairness. Humans, on the other hand, may have differing opinions on whether it is fair to include an input feature in the decision-making process. The authors of \cite{Grgic-HlacaRGW18} suggest a method for figuring out why people think such features are fair or unfair. They introduced seven criteria by which a user assesses a feature with regards to reliability, importance, privacy, volitionality, causes outcome, causes vicious cycle, causes disparity in outcomes, and is caused by sensitive group membership. The purpose of using a combination of classifiers rather than one has motivated us. For example, the authors of \cite{grgic2016case} explored the advantages of replacing a single classifier with a diverse ensemble of random classifiers with regards to accuracy as well as individual and group fairness. In this study, we concentrate on this concept and suggest ${LIME}_{out}$, a tool for ensuring Process fairness while enhancing or at the very least maintaining the model's accuracy.

\section{Methodology}
In this segment, we describe the ${LIME}_{out}$ framework, which consists of two main components: ${LIME}_{global}$ and ${Ensemble}_{out}$. It takes a classifier and a dataset as inputs. The first variable then determines if the classifier is biased on the dataset since the predictions are based on sensitive features. To do this, we make use of ${LIME}_{global}$~\cite{ribeiro2016should}. This will compile a list of the most important features (globally). If sensitive features are among the most important, the classifier is deemed unfair, and the second component of ${LIME}_{out}$ implemented. Other than that, the classifier is deemed fair, and no action is taken. The second component is the fundamental of ${LIME}_{out}$. ${Ensemble}_{out}$ uses feature-drop to generate a pool of classifiers based on the most important features. Each of these classifiers are independent of the corresponding sensitive features they are based on. It then uses this pool of classifiers to construct an ensemble. The choice of sensitive features is left to the user within the given context, following a human and context-centred approach.

\subsection{${LIME}_{global}$}

LIME is widely used to obtain local explanations for instances.  These explanations could be combined to gain some insight into the classifier's global process \cite{ribeiro2016should,van2019global}. To begin,  ${LIME}_{global}$ uses the sub-modular pick method \cite{ribeiro2016should} to select instances. The reliability of the global explanation can be influenced by the instances chosen. The instances obtained from sub-modular pick are used to gain a global understanding of the classifier's inner process. For every one of the instances,  ${LIME}_{global}$ obtains the local explanations (important features and their contributions). As a result, a list of the model's most important features is generated.

\subsection{ ${Ensemble}_{out}$}

${LIME}_{out}$ evaluates the process fairness of any given Machine learning model using the globally important features obtained by ${LIME}_{global}$. We can then check and analyze if the model's predictions are dependent on sensitive features. It is considered unfair or biased if the sensitive features are listed among the top 10 globally important features. In a case where the model is considered to be unfair, one simple solution is to eliminate all sensitive features from the dataset prior to training. These sensitive features, on the other hand, can be strongly correlated with non-sensitive features, maintaining the unwanted bias. also eliminates all such correlated features to minimise this error. This may result in a decline in performance, as the model may become less accurate as a result of the lack of training data after eliminating all the sensitive features. ${LIME}_{out}$ works around this limitation by constructing a pool of classifiers, each of which eliminates a subset of sensitive features. To prevent an exponential number of such classifiers, we only consider those that are obtained by eliminating one or more sensitive features in this study. ${LIME}_{out}$ constructs an ensemble classifier  by combining the pool's classifiers in a linear way. Given an input $(M, D)$, where $M$ is a classifier and $D$ is the dataset. Assume the globally significant features provided by ${LIME}_{global}$ are $a_1,a_2, \ldots, a_n$, with $a_{j1},a_{j2}, \ldots, a_{ji}$, being sensitive. As a result, ${LIME}_{out}$ trains $i + 1$ classifiers: $M_k$ after removing $a_{jk}$ from the dataset, for $k = 1,\ldots,i$, and $M_i$ plus one after removing all sensitive features $a_{j1},a_{j2}, \ldots, a_{ji}$. The ensemble classifier $M_{final}$ is defined as the ``average'' of these $i + 1$ classifiers in this ${LIME}_{out}$ preliminary implementation. More precisely, for an instance $x$ and a class $C$, $P_{M_{final}}(x \in C)=\frac{\sum_{k=1}^{k=i+1}P_{M_{k}}}{i+1}$ As we will see empirically during the experiment over the Russian demographic dataset and set of the classifiers, the dependence of $M_{final}$ on sensitive features decreases, whereas its accuracy is maintained and, in some cases, it even improves.

\section{Experiments}\label{review}

\subsection{Data}
The dataset for the study is obtained from the Research and educational group for Fertility, Family formation and dissolution of the Higher School of Economics. The panel of three waves of the Russian part of Generation and Gender Survey (GGS) was used, which took place in 2004, 2007 and 2011. The dataset contains records of 4857 respondents (1545 men and 3312 women). The gender imbalance of the dataset is caused by the panel nature of the data: the leaving of the survey by the respondents is an uncontrollable process. That is why the representative waves combined in a panel with the structure less close to the general sample.
In the dataset, for each person the following information is indicated: date of birth, gender (male, female), generation, level of education (general, higher, professional), locality (city, town, village), religiosity (yes, no), frequency of church attendance (once a week, several times in a week, minimum once a month, several times in a year or never) and the date of significant events in their lives such as: 1st employment, completion of education of the highest level, leaving the parental home, 1st partnership, 1st marriage, 1st childbirth, breakup of the 1st partnership and divorce of the 1st marriage. Respondents were divided by eleven generations: first (those who were born in 1930--34), second (1935--39), third (1940--44), fourth (1945--49), fifth (1950--54), sixth (1955--59), seventh (1960--64), eighth (1965--69), ninth (1970--74), tenth (1975--79) and eleventh (1980--84)~\cite{IgnatovMMG15}.

\subsection{Experiment set up}
To carry out the experiment, the dataset was split into 70\% to 30\% ratio for training and  test sets, respectively. The original and ensemble models for the dataset were trained using Scikit-learn implementations of the following four algorithms: Logistic Regression, Random Forest, Bagging and AdaBoost. Also, the default parameters of the Scikit-learn documentation was kept for all the algorithms mentioned.	

\subsection{Accuracy Assessment}
In Table 1, we can see the results of the average accuracy obtained in all experiments. For each dataset, the average of the original model’s accuracy and the average accuracy of the ${LIME}_{out}$  ensemble model were indicated. The analysis is based on the comparison between the accuracy of the original and the ensemble models. Since sensitive features were dropped, it was expected that the accuracy of model decreases. However, it is clear that ${LIME}_{out}$  ensemble models maintain the level of accuracy, even though sensitive features were dropped out. We notice a slight improvement in the accuracy of the ensemble models for Bagging in divorce of the first marriage improved from 0.909 to 0.910, Bagging in Religion increased from 0.828 to 0.836, Bagging in Work increased from 0.924 to 0.928, there was also an increase in Random Forest in Work from 0.940 to 0.942, while Bagging in Education also increased from 0.891 to 0.894 over the dataset. Although in one case of Random Forest in divorce, we notice a slight difference between the original and ensemble models the accuracy decreased from 0.918 to 0.914, In any case, the difference is statistically insignificant.

\begin{table}[]
\caption{Average accuracy of assessment}\label{tbl-1}

 \centering
\begin{tabular}{llllll}
\hline\\
\multicolumn{1}{c}{\textbf{Characteristics}} & \multicolumn{1}{c}{\textbf{Approach}}               & \multicolumn{1}{c}{\textbf{LR}}                       & \multicolumn{1}{c}{\textbf{RF}}                       & \multicolumn{1}{c}{\textbf{Bagging}}                  & \multicolumn{1}{c}{\textbf{ADA}}                      \\
\hline\\
Divorce of the 1stmarriage                   & \begin{tabular}[c]{@{}l@{}}Global\\ ${Lime}_{out}$\end{tabular} & \begin{tabular}[c]{@{}l@{}}0.901\\ 0.901\end{tabular} & \begin{tabular}[c]{@{}l@{}}0.918\\ 0.914\end{tabular} & \begin{tabular}[c]{@{}l@{}}0.909\\ 0.910\end{tabular} & \begin{tabular}[c]{@{}l@{}}0.901\\ 0.901\end{tabular} \\
Religiosity                                  & \begin{tabular}[c]{@{}l@{}}Global\\ ${Lime}_{out}$\end{tabular} & \begin{tabular}[c]{@{}l@{}}0.802\\ 0.802\end{tabular} & \begin{tabular}[c]{@{}l@{}}0.841\\ 0.841\end{tabular} & \begin{tabular}[c]{@{}l@{}}0.828\\ 0.836\end{tabular} & \begin{tabular}[c]{@{}l@{}}0.801\\ 0.801\end{tabular} \\
1st employment                               & \begin{tabular}[c]{@{}l@{}}Global\\ ${Lime}_{out}$\end{tabular} & \begin{tabular}[c]{@{}l@{}}0.918\\ 0.918\end{tabular} & \begin{tabular}[c]{@{}l@{}}0.940\\ 0.942\end{tabular} & \begin{tabular}[c]{@{}l@{}}0.924\\ 0.928\end{tabular} & \begin{tabular}[c]{@{}l@{}}0.916\\ 0.916\end{tabular} \\
Completion of education                      & \begin{tabular}[c]{@{}l@{}}Global\\ ${Lime}_{out}$\end{tabular} & \begin{tabular}[c]{@{}l@{}}0.842\\ 0.842\end{tabular} & \begin{tabular}[c]{@{}l@{}}0.902\\ 0.902\end{tabular} & \begin{tabular}[c]{@{}l@{}}0.891\\ 0.894\end{tabular} & \begin{tabular}[c]{@{}l@{}}0.845\\ 0.845\end{tabular}\\
\hline
\end{tabular}
\end{table}

\subsection{Fairness Assessment}
In this segment, we look at the process fairness, the effect of feature dropout and the reliance on sensitive features are analysed. To compute feature contributions and build a list of the most important features, we use ${LIME}_{global}$. Rather than providing the lists of feature contributions for all combinations of the dataset and classifiers, for the dataset, the classifier that provided the highest accuracy was selected, in this case Random Forest was used since it provided the highest accuracy over all other classifiers. As a result, we analyse the  ${LIME}_{global}$ explanations for these various combinations. The most important features for this dataset are listed in Tables 2, 3, 4, and 5. We can see that ${LIME}_{out}$reduces the reliance on sensitive features in some instances. In other words, in the list of most important features, the ensemble models generated by our model have less sensitive features. In addition, LIME explanations indicate that the remaining sensitive features (those in the ensemble model's list) contributed less to the overall prediction than the original model.

\begin{table}[]
\caption{LIME explanation of Random Forest on the 1st divorce}\label{tbl-2}
\begin{tabular}{llllll}
\hline
\multicolumn{2}{c}{\textit{\textbf{Original}}}                  & \multicolumn{1}{c}{\textit{\textbf{}}} & \multicolumn{2}{c}{\textit{\textbf{Ensemble}}}                  & \multicolumn{1}{c}{\textbf{}} \\ \hline
\textit{\textbf{Features}}    & \textit{\textbf{Contributions}} & \textit{\textbf{}}                     & \textit{\textbf{Features}}    & \textit{\textbf{Contributions}} &                               \\
church attendance             & -0.232655                       &                                        & church attendance             & -0.235342                       &                               \\
educational level             & -0.032960                       &                                        & breakup of the 1stpartnership & 0.058290                        &                               \\
breakup of the 1stpartnership & 0.023098                        &                                        & educational level             & -0.051764                       &                               \\
locality                      & 0.020228                        &                                        & locality                      & 0.029759                        &                               \\
1st partnership               & -0.016584                       &                                        & divorce of the 1st marriage   & -0.023312                       &                               \\
gender                        & 0.012340                        &                                        & generation                    & 0.022107                        &                               \\
generation                    & 0.009567                        &                                        & 1st partnership               & -0.006475                       &                               \\
divorce of the 1st marriage   & -0.007100                       &                                        & childbirth                    & -0.002652                       &                               \\
childbirth                    & 0.003008                        &                                        & gender                        & 0.001710                        &                               \\
leaving parents               & 0.0                             &                                        & 1st marriage                  & 0.0                             &                               \\ \hline
\end{tabular}
\end{table}

\begin{table}[]
\caption{LIME explanation of Random Forest on religiosity}\label{tbl-3}
\begin{tabular}{llllll}
\hline
\multicolumn{2}{c}{\textit{\textbf{Original}}}                  & \multicolumn{1}{c}{\textit{\textbf{}}} & \multicolumn{2}{c}{\textit{\textbf{Ensemble}}}                  & \multicolumn{1}{c}{\textbf{}} \\ \hline
\textit{\textbf{Features}}    & \textit{\textbf{Contributions}} & \textit{\textbf{}}                     & \textit{\textbf{Features}}    & \textit{\textbf{Contributions}} &                               \\
church attendance             & -0.232655                       &                                        & church attendance             & -0.235342                       &                               \\
educational level             & -0.032960                       &                                        & breakup of the 1stpartnership & 0.058290                        &                               \\
breakup of the 1s tpartnership & 0.023098                        &                                        & educational level             & -0.051764                       &                               \\
locality                      & 0.020228                        &                                        & locality                      & 0.029759                        &                               \\
1st partnership               & -0.016584                       &                                        & divorce of the 1st marriage   & -0.023312                       &                               \\
gender                        & 0.012340                        &                                        & generation                    & 0.022107                        &                               \\
generation                    & 0.009567                        &                                        & 1st partnership               & -0.006475                       &                               \\
divorce of the 1st marriage   & -0.007100                       &                                        & childbirth                    & -0.002652                       &                               \\
childbirth                    & 0.003008                        &                                        & gender                        & 0.001710                        &                               \\
leaving parents               & 0.0                             &                                        & 1st marriage                  & 0.0                             &                               \\ \hline
\end{tabular}
\end{table}

\begin{table}[]
\caption{Table 4. LIME explanation of Random Forest on the 1st employment}\label{tbl-4}
\begin{tabular}{llllll}
\hline
\multicolumn{2}{c}{\textit{\textbf{Original}}}                  & \multicolumn{1}{c}{\textit{\textbf{}}} & \multicolumn{2}{c}{\textit{\textbf{Ensemble}}}                  & \multicolumn{1}{c}{\textbf{}} \\ \hline
\textit{\textbf{Features}}    & \textit{\textbf{Contributions}} & \textit{\textbf{}}                     & \textit{\textbf{Features}}    & \textit{\textbf{Contributions}} &                               \\
church attendance             & -0.232655                       &                                        & church attendance             & -0.235342                       &                               \\
educational level             & -0.032960                       &                                        & breakup of the 1st partnership & 0.058290                        &                               \\
breakup of the 1st partnership & 0.023098                        &                                        & educational level             & -0.051764                       &                               \\
locality                      & 0.020228                        &                                        & locality                      & 0.029759                        &                               \\
1st partnership               & -0.016584                       &                                        & divorce of the 1st marriage   & -0.023312                       &                               \\
gender                        & 0.012340                        &                                        & generation                    & 0.022107                        &                               \\
generation                    & 0.009567                        &                                        & 1st partnership               & -0.006475                       &                               \\
divorce of the 1st marriage   & -0.007100                       &                                        & childbirth                    & -0.002652                       &                               \\
childbirth                    & 0.003008                        &                                        & gender                        & 0.001710                        &                               \\
leaving parents               & 0.0                             &                                        & 1st marriage                  & 0.0                             &                               \\ \hline
\end{tabular}
\end{table}

\begin{table}[]
\caption{LIME explanation of Random Forest on completing education}\label{tbl-5}
\begin{tabular}{llllll}
\hline
\multicolumn{2}{c}{\textit{\textbf{Original}}}                  & \multicolumn{1}{c}{\textit{\textbf{}}} & \multicolumn{2}{c}{\textit{\textbf{Ensemble}}}                  & \multicolumn{1}{c}{\textbf{}} \\ \hline
\textit{\textbf{Features}}    & \textit{\textbf{Contributions}} & \textit{\textbf{}}                     & \textit{\textbf{Features}}    & \textit{\textbf{Contributions}} &                               \\
church attendance             & -0.232655                       &                                        & church attendance             & -0.235342                       &                               \\
educational level             & -0.032960                       &                                        & breakup of the 1st partnership & 0.058290                        &                               \\
breakup of the 1stpartnership & 0.023098                        &                                        & educational level             & -0.051764                       &                               \\
locality                      & 0.020228                        &                                        & locality                      & 0.029759                        &                               \\
1st partnership               & -0.016584                       &                                        & divorce of the 1st marriage   & -0.023312                       &                               \\
gender                        & 0.012340                        &                                        & generation                    & 0.022107                        &                               \\
generation                    & 0.009567                        &                                        & 1st partnership               & -0.006475                       &                               \\
divorce of the 1st marriage   & -0.007100                       &                                        & childbirth                    & -0.002652                       &                               \\
childbirth                    & 0.003008                        &                                        & gender                        & 0.001710                        &                               \\
leaving parents               & 0.0                             &                                        & 1st marriage                  & 0.0                             &                               \\ \hline
\end{tabular}
\end{table}

\section{Conclusion}

We demonstrated how to use LIME to assess model fairness for demographic data with ${LIME}_{out}$, a framework that takes a pair $(M, D)$ of a classifier $M$ and a dataset $D$ as input and outputs a classifier ${M}_{final}$, that is less sensitive to sensitive features while maintaining accuracy.

The feasibility and versatility of the simple concept of feature dropout followed by an ensemble approach are demonstrated in this replication study (the source is in~\cite{Alves:2020}). This may give rise to numerous future developments and further study. ${LIME}_{out}$ was experimented on four different types of classifiers, but it can smoothly be implanted to other Machine Learning models and data types, together with different explanatory models. An innovative approach, such as~\cite{van2019global}, can be relevant to achieve a global explanation, and this should be thoroughly explored. In addition, the workflow could be enhanced further, for example, classifier ensembles could take into account classifier weighting and other classifiers resulting from the removal of various subsets of sensitive features. A human-centered, context-centered approach was used in this study, which required domain expertise (a professional demographer was involved) for identifying sensitive features in a given use-case. However, this task may be automated, maybe using a metric or utility-based approach to evaluate sensitivity that integrates domain knowledge. This is an interesting task to consider for in the future.

\subsubsection*{Acknowledgments.}
We would like to thank Anna Muratova and Dmitry Ignatov for their help during the paper preparations. 

This research is supported by the Faculty of Social Sciences, National Research University Higher School of Economics. The reported study was performed under the ``ERA.Net RUS plus'' program (RUS\_ST2019-423 -- LifeTraR) and funded by RFBR, project number 20-511-76006.  The contribution of the first author was also prepared within the framework of the HSE University Basic Research Program and performed in the Laboratory for Models and Methods of Computational Pragmatics. 

\bibliographystyle{splncs.bst}

\bibliography{Ibe}

\end{document}